# Process Monitoring of Extrusion Based 3D Printing via Laser Scanning

M. Faes, F. Vogeler, K. Coppens, H. Valkenaers, E. Ferraris
*KU Leuven, Department Of Mechanical Engineering, Leuven, Belgium*

W. Abbeloos, T. Goedemé
*KU Leuven, Department of Electrical Engineering, Leuven, Belgium*

ABSTRACT: Extrusion based 3D Printing (E3DP) is an Additive Manufacturing (AM) technique that extrudes thermoplastic polymer in order to build up components using a layerwise approach. Hereby, AM typically requires long production times in comparison to mass production processes such as Injection Molding. Failures during the AM process are often only noticed after build completion and frequently lead to part rejection because of dimensional inaccuracy or lack of mechanical performance, resulting in an important loss of time and material. A solution to improve the accuracy and robustness of a manufacturing technology is the integration of sensors to monitor and control process state-variables online. In this way, errors can be rapidly detected and possibly compensated at an early stage. To achieve this, we integrated a modular 2D laser triangulation scanner into an E3DP machine and analyzed feedback signals. A 2D laser triangulation scanner was selected here owing to the very compact size, achievable accuracy and the possibility of capturing geometrical 3D data. Thus, our implemented system is able to provide both quantitative and qualitative information. Also, in this work, first steps towards the development of a quality control loop for E3DP processes are presented and opportunities are discussed.

## 1 INTRODUCTION

### 1.1 Additive Manufacturing

Until a decade ago Additive Manufacturing (AM) was merely used as a production method for creating prototypes [1]. However, a recent survey showed that AM nowadays is increasingly used for the production of functional parts. Moreover, many of these AM parts are used in industries where reliability and quality certification play an important role, proving that AM can be used to create functional products with an high added value [2].

Due to the nature of AM, namely adding material in a layer wise approach, for the production of a component, the process is rather slow as compared to conventional mass production techniques, such as e.g. Injection Molding. On the other hand, AM allows the production of complex geometries, including undercut features without the need for a mold, permitting it to be used for the economical manufacturing of single or small series in the context of mass-customization.

Although most AM technologies have developed into mature production processes, a lot of these processes still produce parts out of specifications (e.g. in terms of dimensional inaccuracy or mechanical performance) or have failures that require a restart of the entire build. This increases the use of resources such as time and material, and eventually the cost of the final part.

Online control is a key aspect into reducing failure rate during production, also contributing to process optimization and part quality certification. Therefore, it is most likely that the next trend in AM is to integrate dedicated sensors able to monitor the process state variables and to report possible errors to machine operators for correction and/or process interruption. Moreover, monitored state variables can directly be fed back to the controller, so that the process can become self-adaptive. Furthermore, recording the build of a part can be useful for certification purposes. Monitoring of the process can also help predicting the process outcome, and detecting possible errors in forehand. This philosophy is best described as "Zero Defect Additive Manufacturing", i.e. aiming at (nearly) zero production failure rate. A possible Zero Defect scheme is presented in Fig. 1.

### 1.2 Extrusion Based 3D Printing

Extrusion-based 3D printing (E3DP) processes build up parts in a layer-by-layer method by extruding viscous material through a nozzle.

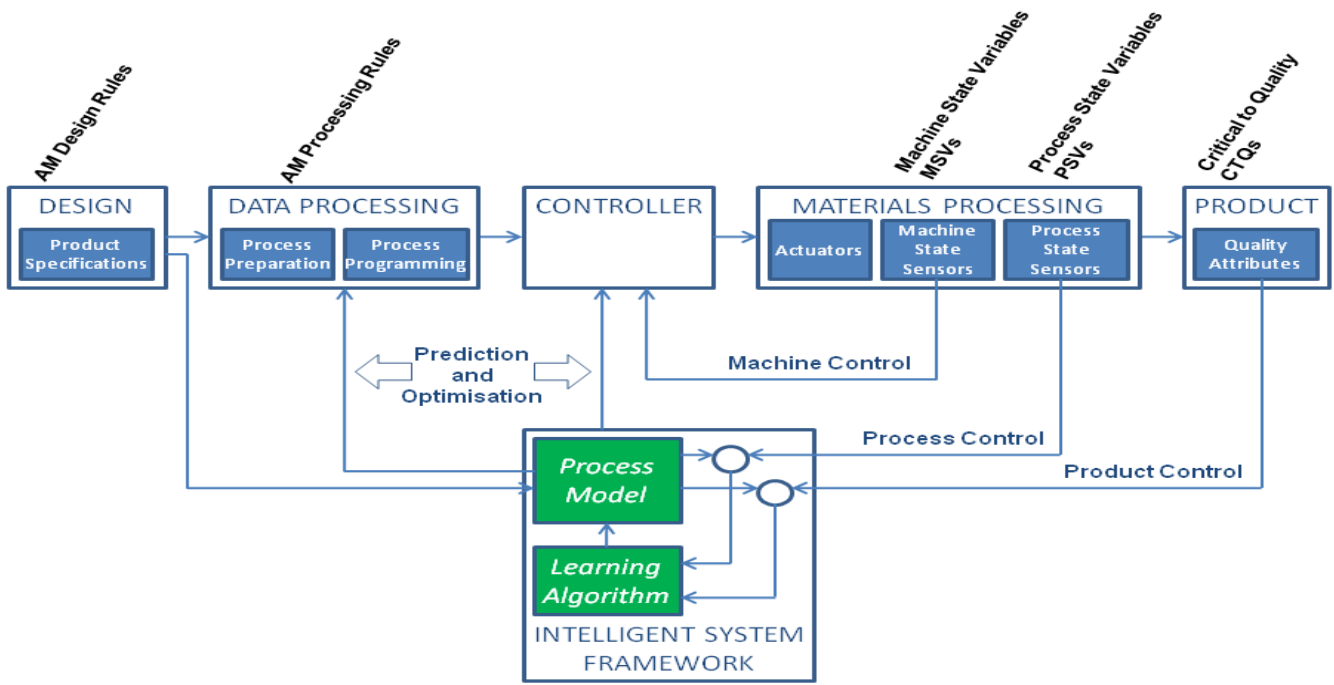

Fig. 1 – Schematic view of a closed loop quality control strategy applicable in the frame of a Zero Defect AM strategy

The achievable resolution in the z-direction (i.e. the layer thickness), is dependent on several parameters, such as deposition velocity, screw rotational velocity, distance between the nozzle and the build platform and the nozzle diameter. Small values, up to 50 microns, can be achieved.

The best-known E3DP technique is Fused Deposition Modelling™ (FDM) [3]. FDM creates components through the track-by-track and layer-by-layer deposition of a heated thermoplastic material. This technique is currently the most applied technique for the production of plastic prototypes [2]. An important drawback of FDM is the limited range of workable materials due to the applied technology (filament extrusion) and current structure of the market (supplier monopoly). The use of a filament also entails an extra production step.

Recently, several screw extrusion based systems have been developed [4–7]. Through the usage of a tailored extrusion screw, a theoretically unlimited range of thermoplastics, blends and polymer composites is workable. The main disadvantage of these systems however is the more complex design and increased cost of the printing head.

Work on online process monitoring and control of Extrusion Based AM techniques is currently limited. However recent research initiatives have been taken to control laser-based metal AM-techniques [8–11] using vision based systems (IR camera's or 3D sensors). Camera based monitoring systems are also already widely used in various industrial branches. Recent work with applications in Extrusion Based 3D-Printing is briefly summarized hereunder.

Fang et al. (2001) developed, implemented and tested an online surface quality monitoring system for parts, produced with FDM of ceramic materials, using signature analysis technique. Image grayscales of a layer were taken as the representative signature that is to be analyzed. Therefore, this image is combined with the tool path that is sent to the printer. Changes in the grayscale values, extracted from various regions on the surface of a layer, enable the possibility to detect under- and overfills on a layer and correct these during the production process [12].

Recently, Heralić et al. (2010) successfully applied a 2D laser triangulation optical measuring system in Laser Metal Wire Deposition, in order to determine the dimensions of the deposited tracks, significantly improving process robustness [13].

At last, Dinwiddie et al. (2013) used an extended-range IR camera to make online temperature measurements of samples made with the FDM process, in order to understand the influence of variations in temperature of the surroundings on variations in mechanical properties of these parts. Temperature differences up to 28ºC in their test set-up, leading to thermal stresses and –distortions [14].

1.3 *Zero Defect Additive Manufacturing*

Due to the specific application of AM techniques, i.e. the production of single or small series, the application of conventional statistical process control strategies is of limited relevance. Therefore, alternative ways for control, monitoring and process prediction, as for targeting (nearly) zero production failure rate (Zero defect AM concept (Fig. 1)) are highly desirable.

The input for any E3DP process (and AM by extension) is a digital CAD-model of the part, usually a triangulated mesh in the ".stl" (Standard Tessellation Language) format [15]. From this file, processing variables, such as extrusion paths, temperature settings, movement speeds, etc. are determined using a dedicated protocol. These settings are sent to the controller, responsible for driving the machine actuators. In conventional CNC machines, machine state variables, such as positions or speed of the axis, are fed back to the controller in order to constitute a high accuracy through 'closed loop control'. Most E3DP machines however lack this sort of feedback.

The presented research focuses on the development, implementation and validation of a monitoring system that is capable of measuring significant machine state variables online. These variables can either directly be fed back to the controller in order to constitute closed-loop feedback (compensation/regulation mode), and/or be used to predict control the process output via an intelligent system framework (forward mode). Such an intelligent system framework uses both online and offline data coming from the semi-finished part, the CAD file or experience to model the process and to make predictions for the machine controller. Due to its general approach, the presented scheme (Fig. 1) can be implemented in a Zero Defect strategy for most AM processes. In the presented research a concept for monitoring the process state variables of the E3DP process is demonstrated and validated.

## 2 SENSOR DESIGN AND DEVELOPMENT

### 2.1 Sensor development

The process state variable of the largest interest is, next to the temperature of the material, the cross-sectional shape and -dimensions of the extruded track, which can be assumed to be elliptic [16]. The sensor should thus be capable of determining the thickness and width of the track, while being as compact as possible.

Moreover, the sensor should at least have a measurement accuracy of 10μm, in order to have sufficient scanning resolution, taking in account that the width of the extruded tracks measures typically between 100μm and 300μm. A measuring speed of at least 50 frames per second is desirable, in order to be able to take sufficient images of a track, knowing that the print head typically moves at a speed between 10mm/s and 20mm/s. Therefore, a 2D laser triangulation system has been selected for the measurement of the dimensions of the tracks.

In our first set-up, a 2MP USB-microscope with an optical magnification of up to 400x was used in combination with a 650 nm laser with a divagation angle of 0,7mrad in order to implement the laser triangulation system on a low-end E3DP machine for initial testing (Fig. 2).

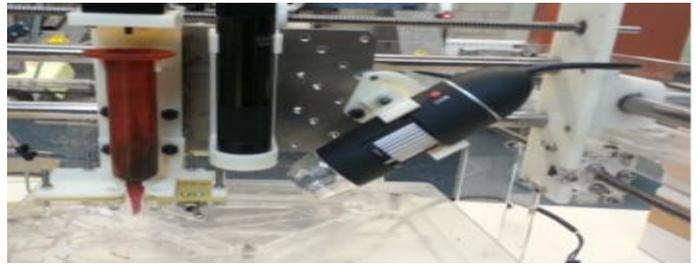

Fig. 2 – Laser Triangulation system implemented on a low-end syringe based E3DP machine

### 2.2 Detection and analysis of the tracks

An algorithm was developed to detect the deposited tracks and determine the dimensions of interest. The algorithm consists of a number of distinct steps.

At first, the laser line is detected by searching for the three maximal intensity values in the extracted image. A parabola is fitted through these points and the maximum hereof is considered the position of the laser line. Secondly, using the detected laser line, the platform is found by taking the median of the pixels, lying on both sides of the image between the edge and $1/8^{th}$ of the width of the image.

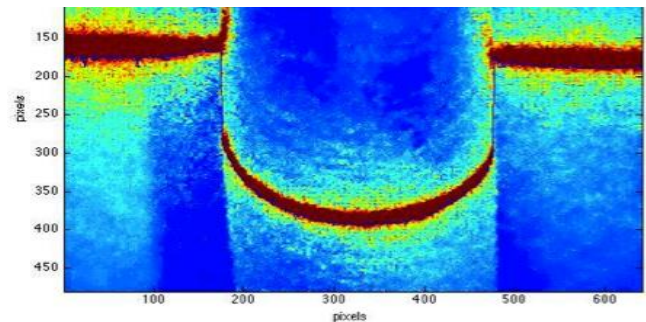

Fig. 3 – Detection of a laser line, projected unto a deposited track of thermoplastic material

At last, in order to detect the track itself, an edge detection algorithm is used on both sides of the track. It is supposed that a side of the track is found when the difference between the laser line and the platform exceeds a certain threshold for 3 subsequent pixels. The height of the track is computed by taking the median of 3 pixels around the center of the track. Using this achieved data, an ellipse is fitted through the extracted points of the laser line and the divergence of the measured points to this ellipse is calculated. The measurement of this divergence is further discussed in section 3.

The triangulation system is calibrated by scanning a rectangular gauge with a thickness of 5mm. It was assumed that all dimensions between 0 and 5mm follow a linear relationship.

### 2.3 Sensor validation

We performed a standardized gage repeatability and reproducibility test in order to determine the performance of the developed triangulation system, using cylindrical calibration pens (3mm-h6).

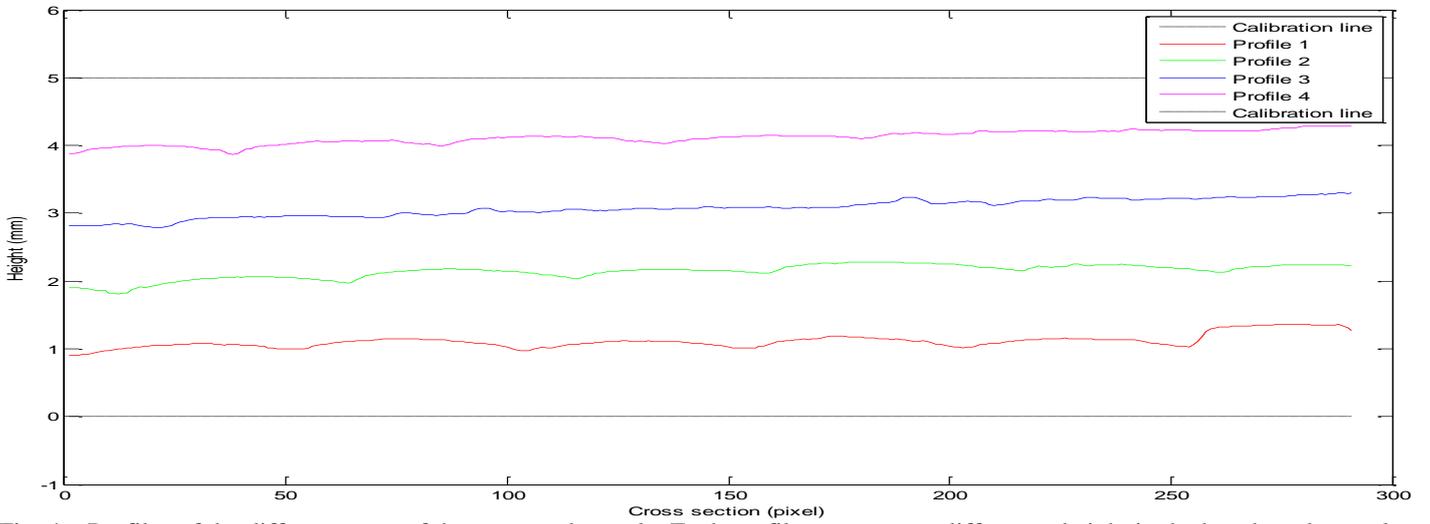

Fig. 4 – Profiles of the different steps of the measured sample. Each profile represents a different z-height in the benchmark sample.

The R&R value of the system was found to be 5pixels at a total of 300 pixels, leading to a variation of 50μm on the measured calibration objects. The main source of this error is the unstable nature of the machine on which the measuring system was implemented. Measurements with an analog gauge revealed a roll of the carrier, supporting the measurement system, of 20μm, being an important addition to the lack of R&R in the triangulation system.

This is not in correspondence with the constraints, discussed in 2.1. Therefore, we used more suitable components, allowing for higher precision, and mounted the entire set-up on a stage with higher stability.

## 3 IMPROVED TRIANGULATION SET-UP

We validated the accuracy of the measurement system against a conventional measuring system. The benchmark sample contained 4 different z-heights (steps), and was produced using FDM. In order to accurately measure the height of each step, calibrated calipers were used. The heights, measured with the laser triangulation system were compared to these measurements. The measurement is shown in Fig. 4, the sample is shown in Fig. 5.

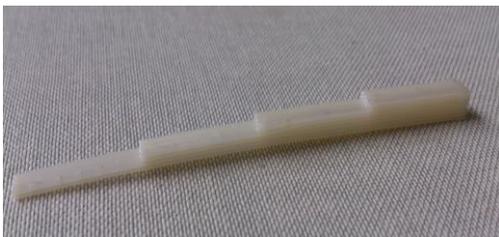

Fig. 5 – Benchmark sample, used for the validation of the improved triangulation set-up

Table 1 gives a comparison between the dimensions, obtained by both systems and the deviation here between. The table demonstrates that a deviation of maximal 100μm is measured, as compared to the calipers.

Table 1 – Comparison of the developed system with a conventional measuring system

|  | Calipers (mm) | Triangulation (mm) | Deviation (mm) |
|---|---|---|---|
| Profile 1 | 1.38 | 1.36 | 0.02 |
| Profile 2 | 2.38 | 2.28 | 0.10 |
| Profile 3 | 3.38 | 3.30 | 0.08 |
| Profile 4 | 4.27 | 4.28 | 0.01 |
| Calibration | 5.00 | 5.00 | 0.00 |

A second source of error in the measurement is the diffusion of the laser light in the polymer material. In order to investigate this, tests were performed on a static setup with improved components. Tests were performed using a 405nm 5mW laser on two common thermoplastic FDM materials (ABS and PLA) in a variety of colors.

The laser power was constant, so the lens aperture was varied to achieve consistent brightness. The extracted points of the laser line were compared to the fitted ellipse and the deviation was assumed to be the error due to the diffusion.

Fig. 6 gives an example of this experiment. The errors, coming from the diffusion, are listed in Table 2. Even though the lowest error was achieved on the dark brown, translucent, PLA sample, we assume the gray ABS material will be the most suitable material to use in real applications. The translucent materials reflect very little light, most of it is refracted, and we expect internal reflections may produce large errors on more complex geometry.

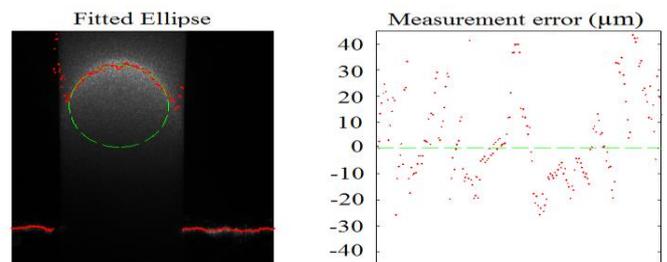

Fig. 6 – Ellipse, fitted through the extracted laser line (L), measurement error as compared to this ellipse (R) of the ABS-white-translucent material

Table 2 – Average error due to the diffusion of the laser light in the polymeric material

| Material | Average error (μm) |
|---|---|
| PLA - red | 8.42 |
| PLA - green - translucent | 7.11 |
| PLA - dark brown - translucent | 4.43 |
| ABS - red | 8.65 |
| ABS - green | 11.20 |
| ABS - gray | 6.15 |
| ABS - white - translucent | 13.86 |
| Mean | 8.55 |

## 4 PROCESS MONITORING AND -CONTROL

The developed measuring set-up will be used to achieve a Zero Defect AM process through the implementation of a closed feedback loop. The goal is to keep the geometrical error in the z-direction of our manufactured part equal or lower than one layer thickness.

A first approach is to continuously vary the layer thickness proportionally to the measured 'z-error'. This can be achieved either by changing the extruder feed rate (g/s) or by changing the movement speed (m/s) of the nozzle online.

If we wish to keep same the extrusion rate and movement speed an alternative approach is to simply add an identical copy of the last layer or skip the next layer as soon as the 'z-error' exceeds half a layer thickness. A more refined variation of this approach is to re-slice the CAD model while printing to achieve the 'perfect' layer geometry, which may differ slightly from the previous layer.

## 5 CONCLUSION AND FUTURE WORK

Extrusion based 3D Printing (E3DP) is an Additive Manufacturing (AM) technique that extrudes thermoplastic polymer in order to build up components using a layer-by-layer approach. Due to this approach, Additive Manufacturing (AM) typically requires long production times in comparison to mass production processes such as Injection Molding. Failures at a certain level during the AM process are often only noticed offline after build completion and frequently lead to the rejection of the part in terms of dimensions and performance, providing an important loss of machine time and material.

A possible solution to improve the accuracy, reliability and robustness of a manufacturing technology is the integration of sensors to monitor and control the process. In this way, errors can be detected online and compensated in an early stage. In this regard, we developed a 2D laser triangulation system, implemented it and performed initial tests. Measurement errors of up to 100μm were found. Sources of error were identified, so that in future work, the system can be improved by implementing it on a more stable stage and selecting more performing components. Finally, we proposed a number of possibilities for online improvement of the z-accuracy of a FDM-made sample. Implementing and comparing the suggested strategies will be addressed in the future.

## ACKNOWLEDGEMENT

Research funded in the framework of the EU-CORNET-IWT project Zero Defect Additive Manufacturing (ZeDAM, EU-Cornet IWT 120280)